\documentclass[lettersize,journal]{IEEEtran}
\usepackage{amsmath,amsfonts}
\usepackage{algorithmic}
\usepackage{algorithm}
\usepackage{array}
\usepackage[caption=false,font=normalsize,labelfont=sf,textfont=sf]{subfig}
\usepackage{textcomp}
\usepackage{stfloats}
\usepackage{url}
\usepackage{verbatim}
\usepackage{graphicx}
\usepackage{cite}
\usepackage{xcolor} %
\usepackage{colortbl,booktabs} %
\usepackage{multirow} %
\usepackage{color} %
\newcommand{\qy}[1]{\textcolor{black}{#1}}
\newcommand{\sq}[1]{\textcolor{black}{#1}}
\newcommand{\sqre}[1]{\textcolor{black}{#1}}
\usepackage[misc]{ifsym}

\begin{document}

\title{Divide and Conquer: Improving Multi-Camera 3D Perception with 2D Semantic-Depth Priors and Input-Dependent Queries}

\author{Qi Song, Qingyong Hu${^{\footnotesize{\textrm{\Letter}}}}$, Chi Zhang, Yongquan Chen, Rui Huang${^{\footnotesize{\textrm{\Letter}}}}$
\thanks{This work was supported in part by Shenzhen Science and Technology Program under grant No. JCYJ20220818103006012 and ZDSYS20211021111415025. \textit{(Corresponding author: Qingyong Hu and Rui Huang.)}}
\thanks{Qi Song, Chi Zhang, Yongquan Chen, and Rui Huang are with School of Science and Engineering, The Chinese University of Hong Kong, Shenzhen, Guangdong, 518172, China. \qy{Qingyong Hu is with the Department of Computer Science, University of Oxford, OX1 3PR, UK.} (e-mail: qisong@link.cuhk.edu.cn; huqingyong15@outlook.com; chizhang1@link.cuhk.edu.cn; yqchen@cuhk.edu.cn; ruihuang@cuhk.edu.cn).}}

\markboth{Journal of \LaTeX\ Class Files,~Vol.~14, No.~8, August~2021}%
{Shell \MakeLowercase{\textit{et al.}}: A Sample Article Using IEEEtran.cls for IEEE Journals}


\maketitle

\begin{figure*}[!t]
\centering
\includegraphics[width=1\textwidth]{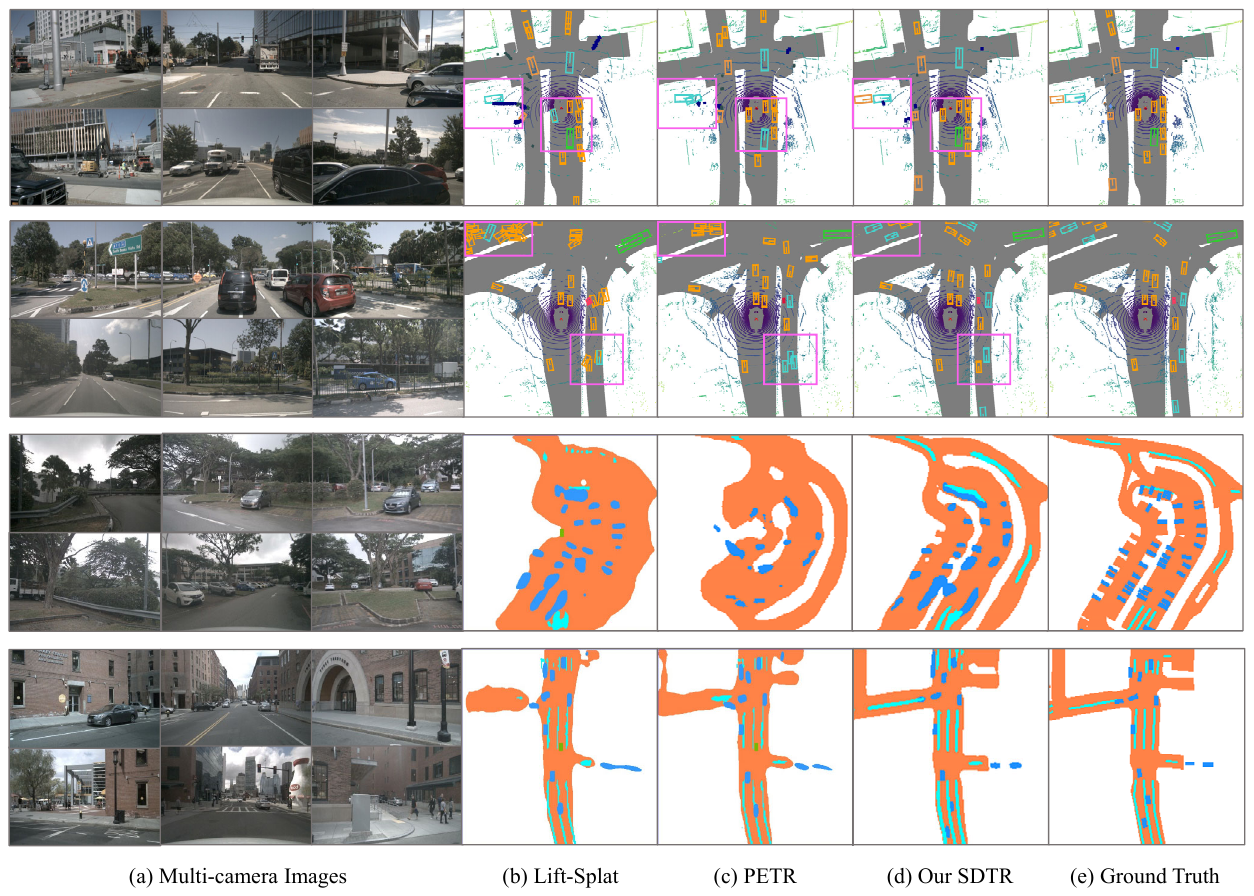} 
\caption{\textbf{Illustration of the multi-camera 3D perception task}. Given the images collected by the cameras from different angles on the vehicle, this task aims to generate segmentation results from the BEV perspective and the 3D object detection results. (a) Multi-camera image inputs. (b) Lift-Splat \cite{philion2020lift} fails to leverage semantic clues in 3D perception, resulting in inaccurate predictions. (c) PETR \cite{Liu2022PETRPE} confuses the spatial arrangement and semantic categories of distant objects without image-specific guidance. (d) Our proposed SDTR model utilizes both semantic-depth priors and input-dependent queries, resulting in significantly improved predictions.}
\label{teaser}
\end{figure*}

\begin{abstract}
3D perception tasks, such as 3D object detection and Bird's-Eye-View (BEV) segmentation using multi-camera images, have drawn significant attention recently. Despite the fact that accurately estimating both semantic and 3D scene layouts are crucial for this task, existing techniques often neglect the synergistic effects of semantic and depth cues, leading to the occurrence of classification and position estimation errors.
Additionally, the input-independent nature of initial queries also limits the learning capacity of Transformer-based models.
To tackle these challenges, we propose an input-aware Transformer framework that leverages Semantics and Depth as priors (named \textbf{SDTR}).
Our approach involves the use of an S-D Encoder that explicitly models semantic and depth priors, thereby disentangling the learning process of object categorization and position estimation. Moreover, we introduce a Prior-guided Query Builder that incorporates the semantic prior into the initial queries of the Transformer, resulting in more effective input-aware queries. 
Extensive experiments on the nuScenes and Lyft benchmarks demonstrate the state-of-the-art performance of our method in both 3D object detection and BEV segmentation tasks.
\end{abstract}


\section{Introduction}
\IEEEPARstart{3}{D} perception is a significant problem in computer vision with diverse applications, including autonomous driving \cite{Caesar2020nuScenesAM} and robot navigation \cite{desouza2002vision}. Of the various tasks involved in 3D perception, multi-camera 3D object detection~\cite{kim2023stereoscopic, xie2023x, Huang2021BEVDetHM, wang2022detr3d} and Bird's-Eye-View (BEV) segmentation~\cite{Schulter2018LearningTL, Mani2020MonoLO, Liu2021WeaklyBD, Liu2020UnderstandingRL, Saha2021EnablingSA} are two representatives that have drawn widespread attention. Despite their different objectives, these tasks aim to infer both \textit{semantic categories} and \textit{3D positions} from 2D cues given by multiple cameras, which makes them ill-posed and entangled with semantic and geometric understanding, presenting significant challenges.

There are two mainstream approaches to multi-camera 3D perception tasks: 1) Depth-based methods, which estimate pseudo-depth to project multi-view image features into 3D space~\cite{philion2020lift,reading2021categorical,Hu2021FIERYFI,zhang2022beverse}. These methods focus on improving depth estimation quality but often overlook the role of semantic cues in reducing classification errors and acting as priors for object localization. Consequently, they tend to underperform, where classification errors and localization errors often appear together, as seen in Figure \ref{teaser} (b). 2) Transformer-based methods, which construct a set of randomly initialized object queries of 3D space~\cite{wang2022detr3d, Liu2022PETRPE} or BEV space ~\cite{li2022bevformer, peng2023bevsegformer} and retrieve relevant image features using a cross-attention mechanism without depth or semantic guidance. However, the input-independent nature of these queries (\textit{i.e.,} all input images share the same object queries) makes training more difficult and reduces detection sensitivity to distant objects, as shown in Figure \ref{teaser} (c).

In this paper, our objective is to develop an effective framework that addresses the challenges outlined above. We believe that \textit{semantics and depth are equally essential to 3D perception but are implicitly learned and tightly coupled in existing networks}, restricting the full exploitation of valuable information. To overcome this limitation, we propose explicitly incorporating semantics and depth as prior knowledge to divide features for classification and position estimation. Additionally, we investigate strategies to make queries input-sensitive for transformer-based methods, alleviating the issues associated with their input-independent nature. Our findings suggest that certain prior knowledge can facilitate achieving this objective.

In particular, we present a transformer-based framework, named SDTR, which models both semantic and depth representations as prior knowledge. Our SDTR consists of two key designs: 1) an S-D Encoder with two branches to reason \textbf{s}emantic and \textbf{d}epth information contained in 2D images with explicit supervision, enabling the network to focus on relevant features and joint objectives; and 2) a Prior-guided Query Builder (\sqre{PQB}) that incorporates image-specific semantic guidance into the initial queries, transforming input-independent queries into input-aware queries, and improving the network's perception capability in complex scenarios. The proposed SDTR is demonstrated to be highly accurate in deducing both semantic categories and 3D positions, as verified in Figure \ref{teaser}(d). Extensive experiments on the nuScenes and Lyft datasets further demonstrate the superiority of our method against other state-of-the-art 3D perception approaches. The main contributions of this paper can be summarized as follows:

\begin{itemize}
\setlength{\itemsep}{0pt}
\setlength{\parsep}{0pt}
\setlength{\parskip}{0pt}
\item We propose SDTR, a transformer-based framework that incorporates semantic and depth priors to improve the capability of inferring both semantic categories and 3D positions.
\item  We introduce an S-D Encoder with explicit supervision to capture both semantic and depth representations, while the Prior-guided Query Builder is designed to encode data-dependent semantic priors and generate input-aware queries.
\item We demonstrate the effectiveness of our framework on two popular 3D perception tasks, including 3D object detection and BEV segmentation, with significant improvements over state-of-the-art methods on the nuScenes and Lyft datasets.
\end{itemize}

\section{Related Work}

\begin{figure*}[h]
  \centering
   \includegraphics[width=1.0\linewidth]{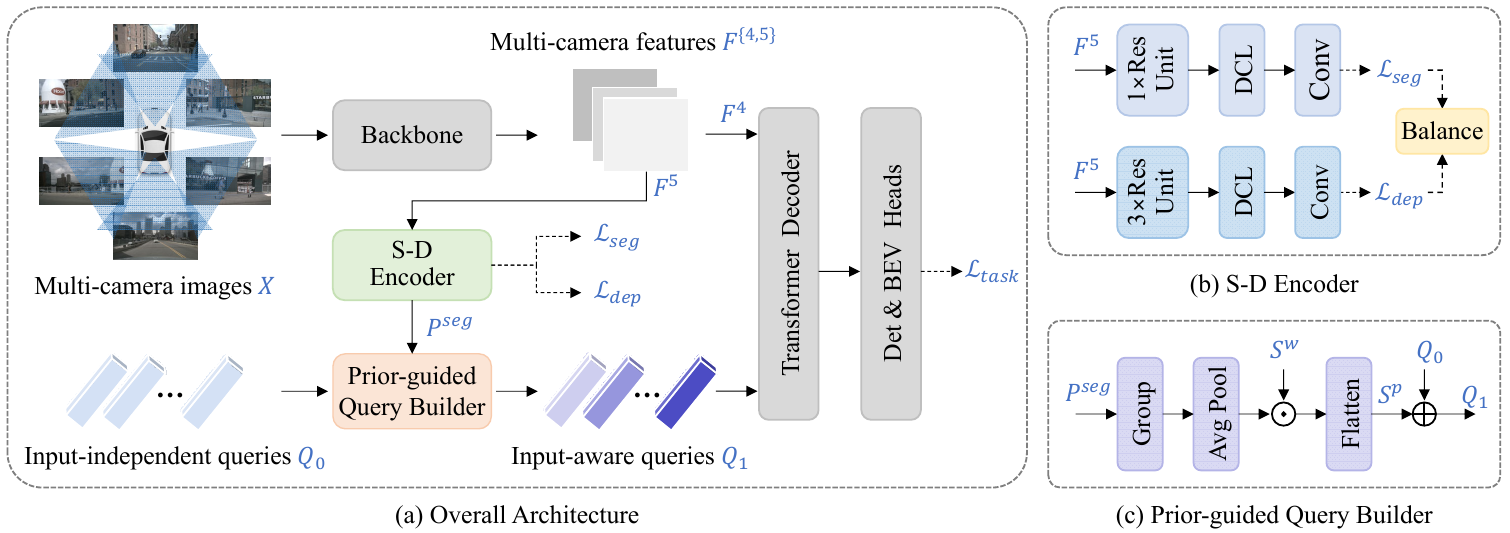}
   \caption{\textbf{Overview of the proposed SDTR framework.} Our model comprises two key components, \textit{i.e.,} the S-D Encoder and the Prior-guided Query Builder, which are designed to effectively extract semantic and depth representations and convert input-independent queries into input-aware queries, respectively. SDTR is capable of producing 3D detection and BEV segmentation results using task-specific heads. \sqre{Specifically, ResUnit and DCL denote the residual unit in \cite{he2016identity} and the dilated convolution layer in \cite{yu2017dilated} respectively.}}
   \label{overall}
\end{figure*}

\subsection{Multi-camera 3D Object Detection}
Multi-camera 3D object detection is a challenging task that involves predicting multi-class 3D bounding boxes from multi-view images. Early work CenterNet \cite{Zhou2019ObjectsAP} predicts 3D properties based on the center point of 2D boxes. Recently, transformer networks \cite{vaswani2017attention, dosovitskiy2020image} have shown promising results in reformulating object detection tasks by constructing a set of object queries and using cross-attention to search for relevant image features. DETR3D \cite{wang2022detr3d}, as a follow-up work of DETR \cite{Carion2020EndtoEndOD}, back-projects the 3D reference points into the image plane to index valid 2D features. PETR \cite{Liu2022PETRPE} perceives the 3D scene information by using the initialized object queries of 3D space. 
\qy{In the wake of rapid advancements in Bird's Eye View (BEV) representation—a favored approach in navigation tasks due to its succinct 2D portrayal of the 3D environment—researchers have put forward to devise a set of BEV queries. These queries facilitate the transformation of perspective between BEV and the image features through cross attention. Both BEVSegFormer~\cite{peng2023bevsegformer} and BEVFormer~\cite{li2022bevformer} employ BEV queries to extract valid features for their ultimate predictions. Nevertheless, the inherent input-independent nature of these queries inadvertently diminishes the detection sensitivity within intricate scenes. To address this, BEVFormerV2 \cite{yang2022bevformer} pioneers a two-stage BEV detector by integrating the first-stage proposals with learnable queries to form the second-stage object queries. However, these queries excessively depend on the precision of high-level 3D detection and continue to exhibit deficiencies in associating input information with learnable queries. To mitigate these limitations, we introduce a novel approach in this paper that involves the incorporation of global semantic priors into the initial queries. This strategy facilitates the generation of input-aware queries, thereby enhancing the flexibility and expressivity of the model.}

\subsection{BEV Segmentation}
BEV segmentation is a task of segmenting objects in the bird's eye view (BEV), which differs from 2D semantic segmentation \cite{Zhao2017PyramidSP, long2015fully,ronneberger2015u, he2022efficient, tian2021prototypical, ke2022three} mainly due to the introduction of perspective shift, leading to ill-posed 2D-to-3D geometry inference. Traditional methods \cite{Lin2012AVB, Abbas2019AGA, Smann2018EfficientSS} use inverse perspective mapping (IPM) to project features from the image plane into the BEV plane. However, IPM only works well in estimating flat road layouts but inevitably introduces errors for 3D objects. To avoid errors introduced by IPM, a handful of works including VED~\cite{Lu2019MonocularSO} and VPN \cite{Pan2020CrossViewSS} directly learn the transformation relation between two planes using the multilayer perceptron. Nevertheless, these approaches damage the spatial information and harms the feature details. Recently, PON \cite{Roddick2020PredictingSM} and PanopticBEV \cite{gosala2022bird} are proposed to use dense transformer layers to map the image features into the BEV space. Lift-Splat \cite{philion2020lift} utilizes the implicit depth distribution to lift multi-view images into 3D coordinates. \sqre{Saha et al. \cite{saha2022pedestrian} introduce a graph neural network to predict BEV objects from monocular images with spatial reasoning.} Another line of work, such as CVT \cite{zhou2022cross} and PYVA \cite{yang2021projecting}, adopts a cross-view transformer to implicitly learn geometric transformation. \qy{In this work, the cross-view transformer is chosen to map perspective view features into bird’s eye view for its strong expressiveness.}

\subsection{Auxiliary Learning for 3D Perception}

\qy{Camera-based 3D perception attempts to understand the semantic layout of the environment and the 3D measurements of objects within it, all from 2D image inputs. This represents a highly intricate and inherently ill-posed learning challenge. In a bid to alleviate convergence difficulties, recent research has been exploring auxiliary tasks as a means of providing comprehensive guidance for the backbone during feature extraction. One prominent avenue of this research utilizes the monocular depth estimation branch \cite{eigen2015predicting, fu2018deep, godard2017unsupervised, godard2019digging, bhat2021adabins} to enhance the ability to interpret 3D geometric understanding from 2D imagery. Notably, BEVStereo \cite{li2022bevstereo} and SOLOFusion \cite{park2022time} exploit depth estimation to lift 2D image features into 3D space. Moreover, BEVDepth \cite{li2022bevdepth} integrates depth supervision to enhance depth prediction capabilities. \sqre{Similarly, the recent work by Dwivedi et al.  \cite{dwivedi2021bird} proposes a novel transformation layer that effectively exploits depth maps to project 2D image features to the BEV space. DAT \cite{zhang2023bev} incorporates depth information to improve the cross-attention mechanism and leverages depth-aware negative suppression loss to prevent duplicate predictions along depth axes.}
A parallel line of research seeks to enhance perception performance through auxiliary image detection. Works such as AutoAlign \cite{chen2022autoalign}, MVX-Net \cite{sindagi2019mvx}, and M\(^{2}\)BEV \cite{xie2022m} incorporate a 2D detection head as an additional training signal. Meanwhile, BEVFormerV2 \cite{yang2022bevformer} constructs a 3D detection head upon the backbone to predict 3D bounding boxes in the perspective view. Furthermore, SimMOD \cite{zhang2022simple} encompasses both 2D detection and 3D key information regression as auxiliary tasks. The commonality in these approaches is the focus on augmenting 3D geometric or 2D detection supervision to refine the backbone features.  However, these methods often overlook the mutually beneficial relationship between the categorical cues and 3D geometry in 2D-to-3D perception. Given that the framework should be adaptable to different 3D perception tasks, e.g., 3D detection, BEV segmentation, etc., this paper opts for a more nuanced semantic supervision, rather than simple 2D detection. In particular, we propose a novel framework that concurrently and comprehensively examines both semantic and depth information with explicit supervision to enhance representation learning. By utilizing semantic and depth priors, our model can more effectively comprehend the 3D scene geometry and provide more precise object segmentation in the BEV space.
}

\section{Method}

\subsection{Overview}

For the multi-camera 3D perception task, \(N\) multi-camera images \(X=\{X_i\in \mathbb{R}^{3\times H_I \times W_I}\}_N\) are given, each with associated extrinsic matrices \(E=\{E_i\in \mathbb{R}^{3\times 4}\}_N\), and intrinsic matrices \(I=\{I_i\in \mathbb{R}^{3\times 3}\}_N\). As depicted in Figure \ref{overall}(a), we first pass the input images through a backbone network to extract multi-view image features, \(F=\{F_i\in \mathbb{R}^{C\times H \times W}\}_N\). In particular, we use the output features of the 4th and 5th stages for subsequent processing. After that, the S-D Encoder is applied on \(F^5\) to jointly predict 2D segmentation map \(P^{seg}=\{P^{seg}_i\in \mathbb{R}^{C_s\times H \times W}\}_N\), and depth map \(P^{dep}=\{P^{dep}_i\in \mathbb{R}^{C_d\times H \times W}\}_N\). The estimated 2D segmentation map \(P^{seg}\) is further utilized to interact with the initial object queries \(Q_0\) in the Prior-guided Query Builder, enabling initial queries with awareness of class-wise semantics. Then, the newly generated queries \(Q_1\), along with the image features \(F^4\), are input to the transformer decoder. Finally, we employ the 3D detection head and the BEV segmentation head separately or jointly for the final prediction. \sqre{In particular, the 3D detection head includes two branches for classification and regression, similar to previous works like DETR3D \cite{wang2022detr3d}, while the BEV segmentation head is comprised of an MLP network followed by a sigmoid layer.}

\begin{figure}[t]
\centering
   \includegraphics[width=1.0\columnwidth]{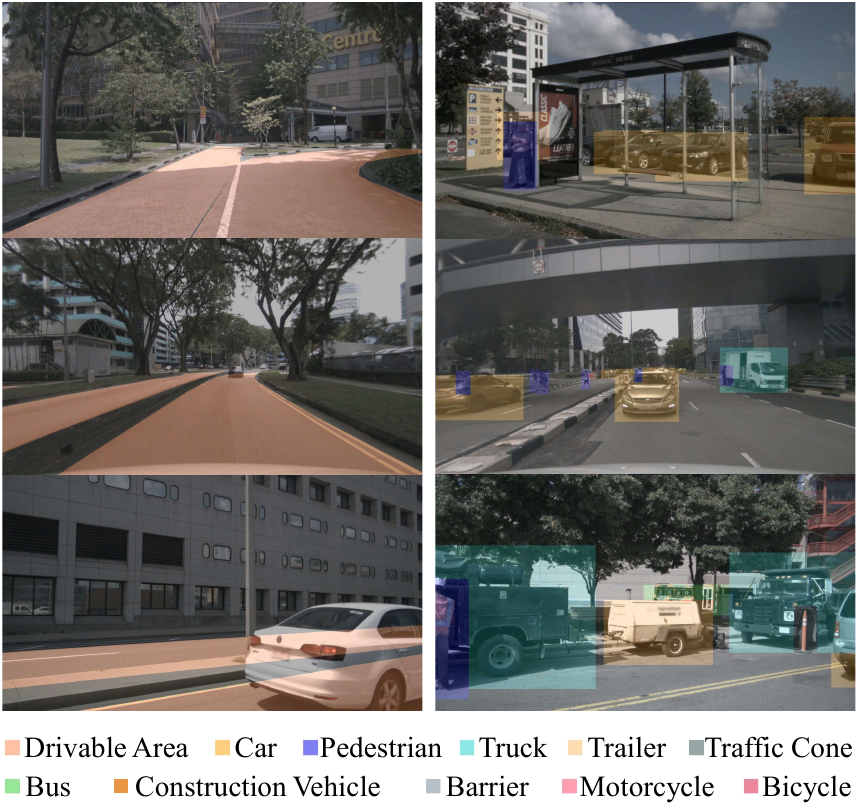}
 
   \caption{\textbf{Visualization of the generated 2D semantic labels.} \qy{The first column exhibits the ground truth of the drivable area, while the subsequent column portrays the object-specific semantic labels. For clear illustration, the semantic labels have been superimposed on the original RGB images, thereby facilitating a more intuitive understanding.}}
   \label{seg_vis}
\end{figure}

\subsection{S-D Encoder}

In the realm of multi-camera 3D perception, it is important to note the fundamental discrepancy between the coordinate systems of input images and output predictions. Moreover, this task necessitates not only the estimation of missing 3D layouts, but also the inference of their corresponding semantic information. Nonetheless, existing methods typically rely on indirect and limited supervision from 3D perception labels, which hinders the network's ability to learn an optimal representation, further exacerbating the difficulty of the task. To address the limitations, we present a semantic-depth joint perception module that incorporates two auxiliary branches to effectively leverage both types of information present in 2D images, with the aim of improving the accuracy and robustness of the 3D perception task. The detailed architecture of our S-D Encoder is shown in Figure \ref{overall}(b).

Considering it is highly challenging for an end-to-end neural network to generate precise depth or semantics, with only indirect and limited supervision from the 3D perception labels, we adopt a disassembled learning process, wherein each branch is explicitly supervised. This enables the accurate learning of semantic and depth features by both branches, which in turn contributes to the 3D perception performance. To acquire 2D semantic labels, we first back-project the BEV labels of the drivable area onto the image plane based on the camera parameters, and then employ the annotations from 2D detection to generate object labels. Figure \ref{seg_vis} shows some examples of the generated 2D semantic labels. \sqre{Within the context of the BEV segmentation task, since 
the segmentation of both road and object elements is required, road labels and object labels are concatenated to form the auxiliary semantic labels. In contrast, for the 3D object detection task, only object labels are utilized, and the segmentation of road elements is not considered.}
As for the depth labels, we utilize the point clouds present in the dataset to derive the ground truth.

On the other hand, to optimize the interplay between semantics and depth in both auxiliary branches, we also endeavored to achieve a balance between them through careful consideration of both architecture design and loss combination. Specifically, given the ill-posed nature of monocular depth estimation, we empirically integrated three \sqre{Residual Units} \cite{he2016identity} into the depth branch, while allocating a single unit to the segmentation branch, taking into account the discrepancy in task complexity. Moreover, we empirically explore various combinations of loss weights to achieve the optimal trade-off, as detailed in Table \ref{ablation2}.

\subsection{Prior-guided Query Builder}

\begin{figure}[t]
\centering
   \includegraphics[width=1.0\columnwidth]{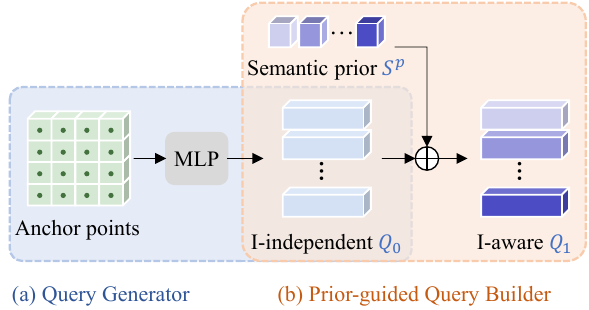}
  
   \caption{\textbf{Comparison of Query Generator in PETR~\cite{Liu2022PETRPE} and our Prior-guided Query Builder.} (a) In PETR, the queries \(Q_0\) are randomly initialized and input-independent. (b) In contrast, SDTR generates input-aware queries \(Q_1\) by encoding image-specific semantic priors, which enhances flexibility and expressiveness.}
   \label{module_compare}
\end{figure}

Existing transformer-based approaches such as PETR~\cite{Liu2022PETRPE} employ a collection of trainable anchor points in 3D space as the initial phase. Despite the fact that encoding 3D space information helps ensure convergence, the initial queries remain randomly initialized and input-independent. This introduces a considerable level of ambiguity to the model learning process and reduces detection sensitivity towards intricate scenarios. To tackle this issue, we propose the integration of data-dependent semantic priors into the initial queries, as illustrated in Figure \ref{module_compare}(b), thereby generating input-aware queries.

\sqre{We have noticed the existence of another method, namely PAP \cite{feng2023priors}, which similarly integrates 2D predictions to formulate query priors. However, the strategy employed by PAP diverges significantly during the processing of these 2D priors. Initially, PAP executes the cropping of depth maps, feature maps, and semantic maps based on the predicted 2D boxes. This approach presumes a concentrated generation of queries centered around individual objects, an assumption that may prove incompatible with the requisites of BEV segmentation. This is because, the BEV segmentation task involves not only object segmentation but also the segmentation of map elements, necessitating a broader scope of query generation. Furthermore, PAP employs multiple 2D cues (such as 2D boxes, semantic maps, and depth maps) to construct the query priors, which may introduce cumulative errors due to the sparseness of depth labels in nuScenes dataset. In contrast, our PQB framework leverages only semantic maps, ensuring awareness of map elements and ensuring higher efficiency.
}

The architecture of our Prior-guided Query Builder is presented in Figure \ref{overall}(c). 
Given that the essence of the transformer architecture lies in the identification of relevant features through cross-attention, it follows that for tasks such as 3D detection or segmentation, the selected feature points ought to be derived from the foreground or meaningful objects within the image. In this context, semantic segmentation can serve as a means of acquiring such effective features as prior information. 
Therefore, the first step of our module involves the integration of semantic priors obtained from valuable multi-view features into the query process, which can be formulated as follows:
\begin{equation}
  S^p = \Psi(P^{seg}, S^w)
  \label{lossfunc10}
\end{equation}

\noindent where \(\Psi(\cdot)\) is a collection of operations that map the 2D segmentation map to semantic priors, \qy{\(S^p\) and \(S^w\) denote semantic priors and class-specific weights, respectively.}

Specifically, given the 2D segmentation map \(P^{seg}\in \mathbb{R}^{N\times C_s\times H \times W}\), where \(N\) and \(C_s\) are the number of multi-view images and semantic classes, \(H\) and \(W\) are the spatial dimensions of the feature. The group operation is utilized to aggregate all the views \(\mathbb{R}^{C_s\times (N\times H \times W)}\). Subsequently, average pooling is utilized to generate the global semantic representations \(\mathbb{R}^{C_s\times (N_q/C_s)}\), where \(N_q\) corresponds to the number of queries. Moreover, due to the dissimilarities in the distribution of categories within the image, the global semantic representations are multiplied by a trainable class-specific weight \(S^{w}\in \mathbb{R}^{C_s}\).  The results are flattened to form the semantic priors \(S^{p}\in \mathbb{R}^{N_q}\).

Finally, we incorporate \(S^{p}\) into the input-independent queries \(Q_0\) to obtain the final input-dependent queries  \(Q_1\):

\begin{equation}
  Q_1 = S^p + Q_0
  \label{lossfunc11}
\end{equation}

The generated semantic priors can filter out the required global semantic representations, which provide image-specific guidance for the initial queries. In our experiments, the inclusion of semantic priors resulted in a significant enhancement of the detection performance, particularly for attribute prediction, with minimal computation overhead. Further discussions are provided in Section \ref{sec:ablation}.

\begin{table*}[htp]
 \caption{\textbf{3D detection results on the nuScenes \textit{val} set.} \qy{To ensure a fair comparison, all the reported models were trained without the incorporation of temporal information.} The symbol {\color{blue}\(\dagger\)} indicates that the model was fine-tuned and tested with test time augmentation. }
\centering
\resizebox{1.0\textwidth}{!}{
\begin{tabular}{rcc|cc|ccccc}
\toprule
Method   & Backbone &Resolution& NDS↑  & mAP↑  & mATE↓ & mASE↓ & mAOE↓ & mAVE↓ & mAAE↓  \\
 \midrule
BEVDepth \cite{li2022bevdepth} &R50&512×1408&0.359& 0.312& \textbf{0.718}& \textbf{0.278}& 0.638& 1.150& 0.334 \\
PETR \cite{Liu2022PETRPE}    & R50 & 512×1408 & 0.367& 0.317& 0.840& 0.280& \textbf{0.616}& 0.954& 0.233 \\
\rowcolor[gray]{0.92} \textbf{SDTR (Ours)}      &   R50    & 512×1408 & \textbf{0.384}  &	 \textbf{0.331}& 0.799&	0.280 &\textbf{0.616}	& \textbf{0.904}	&\textbf{0.212}\\
 \midrule
FCOS3D{\color{blue}\(^\dagger\)} \cite{wang2021fcos3d}      &  R101 & 900×1600 &0.415  & 0.343 &0.725& 0.263 &0.422& 1.292& \textbf{0.153}  \\
DETR3D \cite{wang2022detr3d}   & R101   & 900×1600& 0.425 &0.346& 0.773& 0.268& 0.383& 0.842& 0.216 \\
PGD{\color{blue}\(^\dagger\)} \cite{wang2022probabilistic} & R101 &900×1600 & 0.428  & 0.369 & 0.683 & \textbf{0.260} & 0.439  &1.268 & 0.185 \\
BEVFormer-S \cite{li2022bevformer} &  R101  &900×1600 &0.448  & 0.375 & 0.725 & 0.272  &0.391 & 0.802  &0.200  \\
SimMOD \cite{zhang2022simple} &  R101  &900×1600 &0.455& 0.366& 0.698& 0.264& \textbf{0.340} &\textbf{0.784}& 0.197  \\
Ego3RT\cite{lu2022learning}&  R101  &900×1500 & 0.450 &0.375 &\textbf{0.657}& 0.268& 0.391 &0.850& 0.206  \\
BEVDepth \cite{li2022bevdepth} &R101&512×1408&0.408&0.376 &0.659& 0.267& 0.543& 1.059 &0.335\\
PETR \cite{Liu2022PETRPE}   & R101 &512×1408  &0.441 &0.366& 0.717& 0.267& 0.412& 0.834 &0.190 \\
\rowcolor[gray]{0.92}  \textbf{SDTR (Ours)}       &   R101    & 512×1408 &\textbf{0.462} &\textbf{0.380}& \textbf{0.657}& 0.267& 0.386& 0.806 &0.167 \\
 \midrule
M\(^{2}\)BEV \cite{xie2022m}   & X101 &900×1600  & 0.470 &0.417& 0.647& 0.275& \textbf{0.377}& 0.834 &0.245 \\
DETR3D \cite{wang2022detr3d}   & V2-99   & 900×1600& 0.374  &0.303& 0.860& 0.278& 0.437& 0.967& 0.235 \\
BEVDet \cite{Huang2021BEVDetHM} & Swin-T & 512×1408  &0.417 &0.349& 0.637 &0.269& 0.490& 0.914 &0.268 \\
PETR \cite{Liu2022PETRPE}   & Swin-T &512×1408  &0.431 &0.361 &0.732& 0.273& 0.497& \textbf{0.808} &\textbf{0.185} \\
\rowcolor[gray]{0.92} \textbf{SDTR (Ours)}       &  V2-99 &  512×1408  &  \textbf{0.482} &	\textbf{0.430} & \textbf{0.643}&	\textbf{0.265} &0.406	& 0.830	&0.192\\
\bottomrule
\end{tabular}}

  \label{nu_det_val}
\end{table*}

\subsection{Losses}

Our network is trained by leveraging a combination of losses, comprising task-specific losses \(\mathcal{L}_{task}\), 2D segmentation loss \(\mathcal{L}_{seg}\), and depth estimation loss \(\mathcal{L}_{dep}\) for the auxiliary branches:

\begin{equation}
  \mathcal{L}_{total}=\mathcal{L}_{task} + \gamma_{seg} \mathcal{L}_{seg} + \gamma_{dep} \mathcal{L}_{dep} 
  \label{lossfunc1}
\end{equation}

In our experiments, we explore the tasks of 3D detection and BEV segmentation, where \(\mathcal{L}_{task}\) is set to either \(\mathcal{L}_{det}\) or \(\mathcal{L}_{bev}\) for single-task learning, or the weighted sum of \(\mathcal{L}_{det}\) and \(\mathcal{L}_{bev}\) for multi-task learning. The hyperparameter \(\gamma\)  is determined empirically to balance the auxiliary branches.

For \(\mathcal{L}_{det}\) in 3D object detection, 

\begin{equation}
  \mathcal{L}_{det}=\mathcal{L}_{cls}+\mathcal{L}_{reg}
  \label{lossfunc2}
\end{equation}

\noindent where \(\mathcal{L}_{cls}\) is the focal loss for object classification. \(\mathcal{L}_{reg}\) is the ${L}_1$ loss for regression.

\section{Experimental Setup}

\subsection{Datasets and Metrics}
\subsubsection{Datasets} Our proposed approach is evaluated on two extensively used large-scale autonomous driving datasets: nuScenes \cite{Caesar2020nuScenesAM} and Lyft \cite{kesten2019lyft}. The nuScenes dataset comprises 1000 scenes captured in Boston and Singapore, while the Lyft dataset includes 180 scenes. Both datasets provide images captured from 6 calibrated surround-view cameras and LiDAR scans, enabling us to explicitly supervise semantic and depth estimations. Additionally, every scene provides the extrinsic and intrinsic parameters of the cameras. As the Lyft dataset does not offer a canonical train/val split, we adopt the division settings in FIERY \cite{Hu2021FIERYFI} and re-implement previous methods to ensure a fair comparison.

\subsubsection{Evaluation Metrics} For 3D object detection, we employ the standard evaluation metrics including mean Average Precision (mAP), nuScenes Detection Score (NDS), and five True Positive (TP) metrics: mean Average Translation Error (mATE), mean Average Scale Error (mASE), mean Average Orientation Error (mAOE), mean Average Velocity Error (mAVE), and mean Average Attribute Error (mAAE). The NDS metric is a weighted sum of mAP and five TP metrics. Additionally, in the realm of Bird's Eye View (BEV) segmentation, the widely used Intersection over Union (IoU) metric is adopted as the primary evaluation metric.

\subsection{Implementation Details}

Our implementation is based on the PyTorch framework \cite{Paszke2017AutomaticDI}, and our work is focused on two primary 3D perception tasks: 3D object detection and BEV segmentation. Specifically, we employ the ResNet series \cite{he2016identity} and VoVNetV2 \cite{lee2020centermask} as the backbone networks, generating the output features \(F^{\{4,5\}}\) with \(1/16\) input resolution. The transformer decoder with 6 layers is adopted to constantly update the queries. 

In our experiments, we resize and crop the multi-view images to 512×1408 for use as network inputs. To define the perception ranges, we set the X and Y axes to [-61.2m, 61.2m], and the Z axis to [-10m, 10m]. Our network is trained using the AdamW optimizer \cite{loshchilov2017decoupled} with a weight decay of 1\(e\)-2. The learning rate is initialized to 2\(e\)-4 and decayed using the cosine annealing policy \cite{loshchilov2016sgdr}. It is worth noting that, unlike previous works which are typically trained on Tesla A100 or V100 GPUs, we conduct all of our experiments using 8×2080Ti GPUs. The training process runs for 24 epochs with a batch size of 8.

In our 3D object detection task, we leverage 900 detection queries and employ the Focal Loss \cite{lin2017focal} for object classification, along with the L1 loss for 3D bounding box regression. For BEV segmentation, we conduct experiments using 625 BEV segmentation queries and utilize the weighted cross-entropy loss for supervision on the predicted BEV map. Additionally, for both tasks, we employ the binary cross-entropy loss as an auxiliary loss for both depth estimation and 2D segmentation.

\begin{table*}[htp]
\caption{\textbf{3D detection results on the nuScenes \sqre{\textit{test}} set.} \qy{To ensure a fair comparison, all the reported models were trained without the incorporation of temporal information.} The symbol {\color{blue}\(\dagger\)} denotes that the method uses test time augmentation. The BEVDet, DETR3D, PETR, BEVDepth, BEVFormer, and SDTR are all trained with CBGS \cite{zhu2019class}.}
\centering
\resizebox{1.0\textwidth}{!}{
\begin{tabular}{rcc|cc|ccccc}
\toprule
Method     & Backbone & Resolution & NDS↑  & mAP↑  & mATE↓ & mASE↓ & mAOE↓ & mAVE↓ & mAAE↓ \\
 \midrule
CenterNet \cite{Zhou2019ObjectsAP}&       DLA   &-    & 0.400  & 0.338 & 0.658 & 0.255 & 0.629 & 1.629 & 0.142  \\
Ego3RT\cite{lu2022learning}&  R101  &900×1500 & 0.443& 0.389 &0.599& 0.268 &0.470 &1.169& 0.172  \\
FCOS3D{\color{blue}\(^\dagger\)} \cite{wang2021fcos3d}& R101 & 900×1600 & 0.428  & 0.358 & 0.690 & 0.249 & 0.452 & 1.434 & \textbf{0.124}  \\
PGD{\color{blue}\(^\dagger\)} \cite{wang2022probabilistic} & R101 &   900×1600 & 0.448 & 0.386 & 0.626 & 0.245 & 0.451 & 1.509 & 0.127  \\
PETR \cite{Liu2022PETRPE} & R101  &  900×1600 & 0.455 &0.391& 0.647 &0.251& 0.433& 0.933& 0.143\\
SimMOD \cite{zhang2022simple} &  R101  &900×1600 & 0.464& 0.382& 0.623& 0.252& 0.394& 0.863 &0.132\\
Graph-DETR3D \cite{chen2022graph}&  R101  &900×1600 & 0.472 &0.418& 0.668& 0.250 &0.440 &0.876& 0.139\\
M\(^{2}\)BEV \cite{xie2022m} & X101  &  900×1600 & 0.474& 0.429& 0.583& 0.254& 0.376 &1.053& 0.190\\
BEVDet \cite{Huang2021BEVDetHM} & Swin-S  &768×2112 & 0.463  & 0.398 & 0.556 & \textbf{0.239} & 0.414 & 1.010 & 0.153  \\
PETR \cite{Liu2022PETRPE} & Swin-S   & 768×2112 &0.481 &0.434  &0.641 & 0.248 & 0.437 & 0.894 & 0.143 \\
DD3D{\color{blue}\(^\dagger\)} \cite{park2021pseudo}& V2-99    & - & 0.477 & 0.418 & 0.572 & 0.249 & \textbf{0.368} & 1.014 & \textbf{0.124}  \\
Ego3RT\cite{lu2022learning}&  V2-99  &900×1500 & 0.473& 0.425 &0.549& 0.264& 0.433 &1.014 &0.145  \\
DETR3D \cite{wang2022detr3d} & V2-99    &900×1600   & 0.479  & 0.412 & 0.641 & 0.255 & 0.394 & 0.845 & 0.133  \\
BEVDet \cite{Huang2021BEVDetHM} & V2-99    &  900×1600 & 0.488 &0.424& \textbf{0.524}& 0.242& 0.373 &0.950 &0.148 \\
SimMOD \cite{zhang2022simple} &  V2-99  &900×1600 &0.494 &0.417& 0.570 &0.248 &0.387& 0.813& 0.126 \\
Graph-DETR3D \cite{chen2022graph}&  V2-99  &900×1600 &0.495& 0.425& 0.621& 0.251 &0.386& \textbf{0.790} &0.128 \\
BEVFormer-S \cite{li2022bevformer}& V2-99    &900×1600  &0.495   & 0.435& 0.589 &0.254& 0.402& 0.842& 0.131 \\
PETR \cite{Liu2022PETRPE} & V2-99   & 900×1600 &0.504& 0.441 &0.593& 0.249& 0.383& 0.808 &0.132 \\
PETR \cite{Liu2022PETRPE} & V2-99   & 512×1408 &0.495& 0.437 &0.601& 0.248& 0.405& 0.841 &0.142 \\
 \midrule
\rowcolor[gray]{0.92} \textbf{SDTR (Ours)}       &  V2-99 &  \textbf{512×1408}  &  \textbf{0.505} &\textbf{0.449}	 & 0.579&	0.250 &	0.392& 0.833	&0.140\\
\bottomrule
\end{tabular}}
  \label{nu_det_test}
\end{table*}

\begin{table}[t]

 \caption{\textbf{BEV segmentation results on the nuScenes \textit{val} set.} \sqre{Please note that the top section employed different BEV grid settings or validation splits, while the middle section focused exclusively on single-class segmentation, as opposed to the multi-class segmentation studies we utilized. Both sections are included solely for reference purposes. To ensure fairness, we further categorized the remaining methods into two groups based on the application of auxiliary supervision.} {\color{magenta}\(\star\)} denotes that temporal information is utilized. {\color{red}\(^\dagger\)} represents that graph neural network is employed.}
\centering
\small
\begin{tabular}{@{}r|ccc}
\toprule
Method         & IoU-Drive↑ & IoU-Lane↑ &  IoU-Veh.↑\\ \midrule
VED \cite{Lu2019MonocularSO}           & 0.547         & -  & 0.088         \\
VPN \cite{Pan2020CrossViewSS}             & 0.580         & -   &0.255     \\
PON \cite{Roddick2020PredictingSM}               & 0.604         & - & 0.247    \\
LSF\cite{dwivedi2021bird}      &0.611       &     -        & 0.378\\
FISHING \cite{Hendy2020FISHINGNF}          &  - &- & 0.300 \\
STA{\color{magenta}\(^\star\)} \cite{Saha2021EnablingSA}              & 0.707& -  &0.360\\
Image2Map{\color{magenta}\(^\star\)} \cite{saha2022translating}      &0.745&  -           &0.397\\
Ego3RT\cite{lu2022learning}& 0.796&0.475&- \\ \midrule[0.1pt]\midrule[0.1pt]
OFT \cite{Roddick2019OrthographicFT}           & 0.717         & 0.181 & 0.301       \\
Lift-Splat \cite{philion2020lift}   & 0.729  & 0.199& 0.321  \\
FIERY-S \cite{Hu2021FIERYFI}      &-&-& 0.358  \\
FIERY{\color{magenta}\(^\star\)} \cite{Hu2021FIERYFI}      &-&-& 0.382 \\
PedLam{\color{red}\(^\dagger\)} \cite{saha2022pedestrian}      &0.814&  -           &0.498\\ \midrule[0.1pt]\midrule[0.1pt]
BEVFormer-S \cite{li2022bevformer}     &0.807 &0.213 & 0.432 \\
BEVFormer{\color{magenta}\(^\star\)} \cite{li2022bevformer}     &0.801 &0.257 & 0.448 \\  
\rowcolor[gray]{0.92} M\(^{2}\)BEV \cite{xie2022m} &0.759& 0.380 &- \\
\rowcolor[gray]{0.92} \textbf{SDTR (Ours)}   &    \textbf{0.841}     & \textbf{0.476}        &   \textbf{0.450}  \\
\bottomrule
\end{tabular}

  \label{seg_results}
\end{table}

\section{Experimental Results}

\begin{figure*}[htp]
\centering 
   \includegraphics[width=0.82\linewidth]{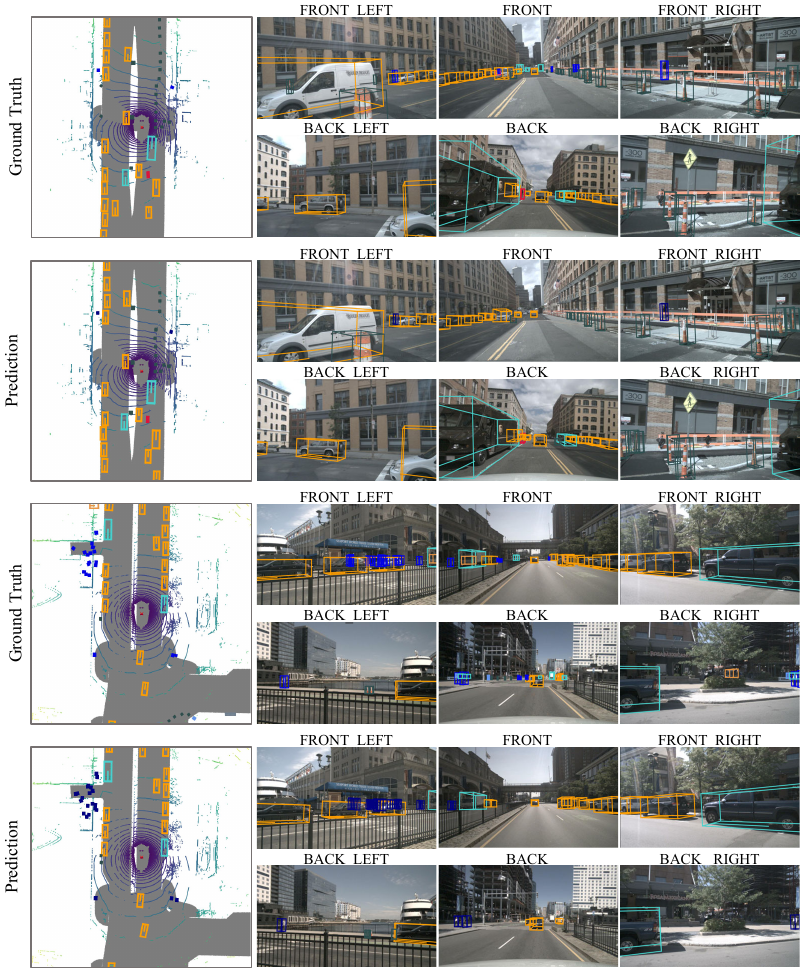}
   \caption{\sq{\textbf{Visualization results for 3D object detection.} We show the 3D bounding box predictions in multi-camera images and the bird’s-eye-view. The 3D bounding boxes are drawn with different colors to distinguish different classes.}}
   \label{quantivative_det}
\end{figure*}

\subsection{State-of-the-art Comparisons}

\qy{When comparing the proposed SDTR with other state-of-the-art (SOTA) methods, it is observed that the majority of these SOTA approaches rely heavily on full—or even supra-maximal—resolution to achieve superior results. Interestingly, the performance enhancements often derive more from the use of high-definition images rather than intrinsic improvements in the methodology itself. 
Furthermore, prior studies typically utilize a large batch size (e.g., 32) \sqre{\cite{li2022bevdepth, Huang2021BEVDetHM, wang2021fcos3d, wang2022probabilistic, lu2022learning}} to ensure the convergence of the model. This requirement inevitably leads to these models being trained solely on high-capacity GPUs such as A100/V100, significantly constraining the practical applicability of the models.
In view of these observations, the experiments in this study have been primarily conducted at a lower resolution (specifically, 512×1408) and with a reduced batch size of 8. These adjustments aim to establish a more equitable basis for comparison and enhance the generalizability of the model.}

\begin{figure*}[t]
\centering
   \includegraphics[width=0.82\linewidth]{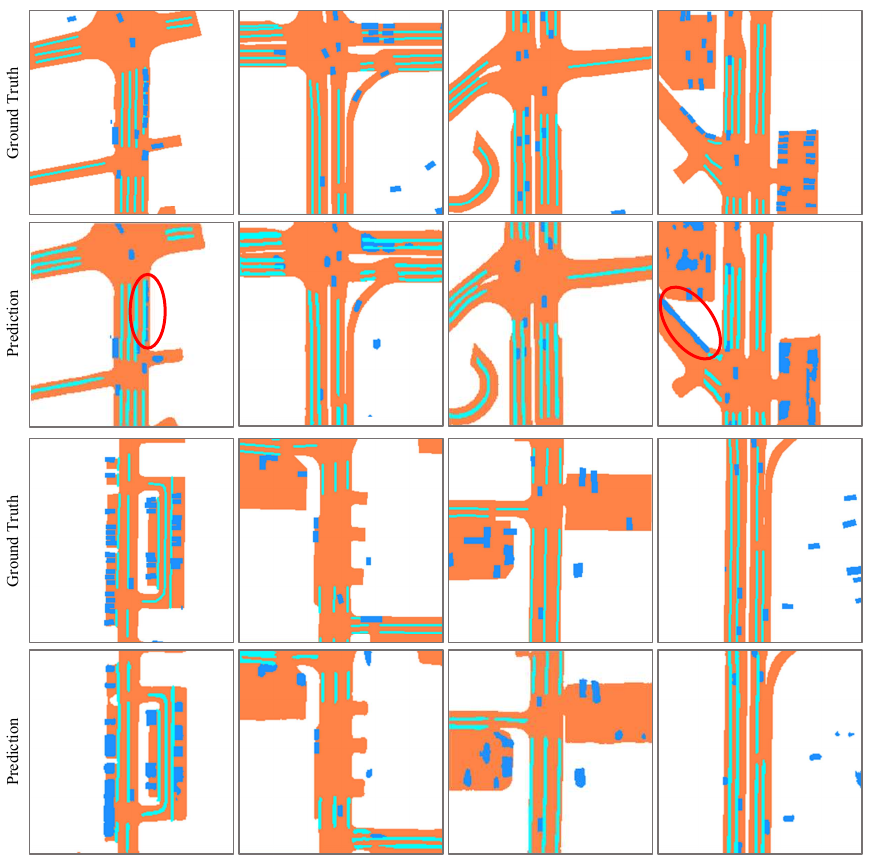}
   \caption{\textbf{Visualization results for BEV segmentation.} Classes of vehicle, drivable area, and lane segmentation are filled with blue, orange, and cyan, respectively.}
   \label{quantivative}
\end{figure*}

\begin{figure}[t]
\centering
   \includegraphics[width=0.98\columnwidth]{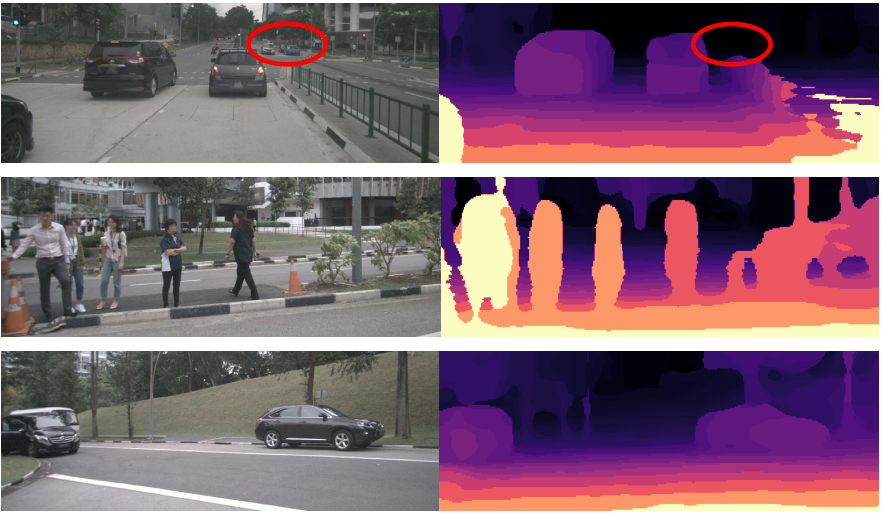}
   \caption{\sqre{\textbf{Visualization of the depth predictions.}} }
   \label{depth_vis}
\end{figure}

\subsubsection{3D Object Detection} We evaluate the performance of our proposed method and other state-of-the-art approaches on the \textit{val} and \sqre{\textit{test}} splits of the nuScenes dataset, as presented in Table \ref{nu_det_val} and Table \ref{nu_det_test}, respectively. On the \textit{val} set, the proposed SDTR outperforms existing paradigms in terms of both NDS and mAP metrics 
\qy{across various existing paradigms, and notably, it is achieved with a smaller input resolution.}
Notably, our model achieves superior performance in accurately reasoning about the 3D scene geometry and semantic category by utilizing both semantic and depth priors, surpassing existing depth-based (e.g., BEVDepth and M\(^{2}\)BEV) and transformer-based (e.g., PETR) approaches. On the \sqre{\textit{test}} set, SDTR also surpasses the previous best method with higher scores of 50.5\% NDS and 44.9\% mAP, while utilizing \textit{only \textbf{1/2} of the input size} compared to its competitors.

\begin{table}[htp]
\caption{\textbf{BEV segmentation results on the Lyft dataset.} \qy{Due to the fact that previous studies employed different validation splits for the Lyft dataset, we have re-implemented these methods in the interest of ensuring a fair comparison.}}

\centering
\small
\begin{tabular}{@{}r|cc@{}}
\toprule
Method       & IoU-Car↑   & IoU-Vehicle↑ \\ \midrule
Lift-Splat \cite{philion2020lift}   &0.389 &  0.382 \\
FIERY-S \cite{Hu2021FIERYFI}        &   -   &  0.410 \\
\rowcolor[gray]{0.92} \textbf{SDTR (Ours)}     &    \textbf{0.457}   &  \textbf{0.451}  \\ \bottomrule
\end{tabular}
  \label{lyft}
\end{table}

\subsubsection{BEV Segmentation} We further provide a comparative analysis of our proposed method against previous state-of-the-art BEV segmentation approaches on both the nuScenes dataset and the Lyft dataset. The results are presented in Table \ref{seg_results} and Table \ref{lyft}, respectively. It is worth noting that our approach consistently outperforms all existing methods and sets a new state-of-the-art performance across all categories. While one prior work, BEVFormer \cite{li2022bevformer}, achieves a high IoU score on the nuScenes dataset by leveraging temporal information and taking full-resolution images as input, our method still achieves better results even in the absence of temporal clues and using smaller input size of 512x1408.

\subsubsection{Visualization Results} Figure \ref{teaser} showcases a visual comparison between our proposed SDTR model and two classical methods in both 3D object detection and BEV segmentation. Owing to the semantic-depth priors and the input-dependent queries, our SDTR model exhibits strong 3D detection performance while ensuring consistent segmentation.
Additionally, Figure \ref{quantivative_det} and Figure \ref{quantivative} provide further qualitative results for 3D object detection and BEV segmentation respectively. These visualizations underscore the proficiency of our SDTR model in executing 3D object detection tasks across diverse scales and distances, as well as BEV segmentation tasks—even under challenging conditions when objects present irregular or extreme shapes.
Despite these promising results, there are instances where our methodology encounters difficulties. Notably, our model may struggle in scenarios where vehicles are densely clustered, as illustrated by the red circles in Figure \ref{quantivative}. Such failure cases are areas of focus for future improvements in our model's performance.

\sqre{Furthermore, as depicted in Figure \ref{depth_vis}, the depth branch integrated within our S-D Encoder demonstrates the capability to accurately estimate depth values, offering valuable 3D positional priors for 2D-to-3D reasoning. However, as indicated by the red circles, there is a tendency for inaccurate depth detection when dealing with distant objects, primarily due to the sparse nature of depth supervision. By incorporating dense semantic priors into our initial queries, we can greatly enhance the recognition of distant classes, and this improvement will be thoroughly demonstrated in Table \ref{distant}. }

\begin{table*}[t]

	\begin{minipage}[b]{0.7\textwidth}
 \caption{\textbf{Ablation studies of proposed modules.} “2D Seg” and “DE” represent the 2D segmentation and depth estimation branches in the S-D Encoder, respectively. We measure the increased computation complexity (measured by the number of FLOPs) and GPU memory usage introduced by the proposed modules without counting the complexity from the baseline. The FPS (frames per second) is measured on a single 2080ti GPU.}\smallskip
		\centering\small
        \tabcolsep=1.5mm
         \begin{tabular}{ccc|ccccc|ccc}
        \toprule
        2D Seg   & DE & PQB& NDS↑& mAP↑  & mATE↓ &  mAOE↓ & mAAE↓& FLOPs& Memory &FPS  \\
         \midrule
        &&&0.360 &  0.312        &      0.834       &       0.659  & 0.237 &       -    &  - &  5.1     \\
        \(\checkmark\)&&&0.367 &   0.321       &      0.816       &        0.644 & 0.230&       30.7G  & 38M  & 5.0      \\
         \(\checkmark\)&&\(\checkmark\)&  0.373&           0.326&              0.810&       0.628&  0.221&     30.7G  & 38M  & 5.0       \\
        \(\checkmark\)&\(\checkmark\)&& 0.377&      0.327    &      0.807       &     0.617  &0.237 &      138.2G   &   648M& 4.6      \\
        \(\checkmark\)&\(\checkmark\)&\(\checkmark\)& 0.384&    0.331      &     0.799           &    0.616 & 0.212    &    138.2G     &  648M & 4.6\\
        \bottomrule
        \end{tabular}
        \label{ablation1}
        
	\end{minipage}
 \hfill
	\begin{minipage}[b]{0.28\textwidth}
\caption{\textbf{Ablation studies of loss combinations.} We adopt different loss weight combinations to directly reflect the effects of auxiliary supervision.}\smallskip
        \centering\small
        \tabcolsep=1.5mm
        \begin{tabular}{cc|cc}
        \toprule
        \(\gamma_{seg}\)        & \(\gamma_{dep}\)         & NDS↑            & mAP↑      \\
         \midrule
        1.0           & 1.0           &     0.378         &    0.329           \\
        2.0           & 1.0           &     0.380           &       0.331           \\
        3.0           & 1.0           &     \textbf{0.384}        &       \textbf{0.331}              \\
        4.0           & 1.0           &     0.379         &   0.328        \\
        \bottomrule    
        \end{tabular}
        \label{ablation2}
        
	\end{minipage}
	
\end{table*}

\begin{table*}[htp]
\caption{\sqre{\textbf{Intersection over Union scores of object categories.} V2-99 is utilized as the backbone network.}}
\centering
\small
\begin{tabular}{ccccccccccc|c}
\toprule
     & Car & Truck & Construction vehicle & Bus & Trailer & Barrier & Motorcycle & Bicycle & Pedestrian & Traffic cone & mIoU\\
 \midrule
SDTR &  0.455   &  0.360     &    0.132          &  0.453   &   0.332      &  0.330       &      0.223      &   0.210      &     0.208       &  0.222  &0.292\\
\bottomrule           
\end{tabular}
  \label{per_iou}
\end{table*}

\subsection{Ablation Studies}
\label{sec:ablation}

To further verify the effectiveness of the proposed modules, we conduct ablation studies from six aspects. All experiments are conducted on the nuScenes \textit{val} set \sqre{and R50 is utilized as the backbone network if not specified.}

\begin{table}[]
\caption{\textbf{Analysis on multi-task joint learning.} We apply multiple loss weight combinations to explore the effect of joint learning, where V2-99 is utilized as the backbone network.}
\centering
\resizebox{\columnwidth}{!}{
\begin{tabular}{cc|cc|ccc}
\toprule
\multicolumn{2}{c|}{Task Head} & \multicolumn{2}{c|}{3D Detection} & \multicolumn{3}{c}{BEV Segmentation} \\
Det           & BEV           & NDS↑            & mAP↑      & Drive↑     & Lane↑           & Vehicle↑   \\
 \midrule
1.0           & 0.0           & \textbf{0.465}                &  \textbf{0.417}              &         -     &  -          &  -        \\
0.0           & 1.0           &       -          &       -         &      \textbf{0.827}        &    \textbf{0.427}        &  0.421   \\
1.0           & 1.0           & 0.462                & 0.412               & 0.815             & 0.415           & 0.413         \\
1.0           & 2.0           &       0.453          &        0.399        &     0.820         &    0.424        &  \textbf{0.423}   \\
\bottomrule    
\end{tabular}}
  \label{multi_task}
\end{table}

\subsubsection{Effectiveness of Proposed Modules} To evaluate the effectiveness of each component in our proposed method, we began with a baseline network and incrementally added the proposed modules. Table \ref{ablation1} summarizes our results. First, it can be seen that the baseline network yielded a NDS and mAP score of 36.0\% and 31.2\%, respectively. Adding the 2D segmentation (2D Seg) branch led to an improvement in NDS and mAP (0.7\% and 0.9\%, respectively), with a negligible increase in computational cost. Our results also indicate that the 2D Seg branch reduced classification and position estimation errors (mAAE, mATE, and mAOE), emphasizing the role of semantic clues as priors for both tasks. Further, combining the 2D Seg and depth estimation (DE) branches led to a more substantial improvement in NDS and mAP (1.7\% and 1.5\%, respectively), demonstrating the effectiveness of decoupling 2D segmentation and depth estimation from 2D-to-3D transformation. 
Given the inherently ill-posed challenge of monocular depth estimation, we empirically utilize a larger quantity of \sqre{Residual Units} in the depth branch. As a consequence, the depth estimation branch necessitated higher computational resources compared to the 2D segmentation branch.
Notably, the integration of \sqre{PQB} allowed our SDTR to achieve superior overall performance, achieving a NDS of 38.4\% and a mAP of 33.1\%. Remarkably, this performance enhancement was achieved without necessitating any additional computational or memory expenditures. \sqre{The reason for constant computational complexity in our PQB is primarily due to the fact that the only trainable parameter it utilizes is a class-specific weight \(S^{w}\in \mathbb{R}^{C_s}\), where \(C_s=\)10 or 3 in 3D detection and BEV segmentation respectively. And such a trainable parameter is significantly smaller in size compared to even the simplest 1x1 convolution layer. Consequently, the increased computational complexity introduced by the PQB is too trivial to be measured or quantified.} These findings suggest that \sqre{PQB} can provide effective guidance for query updating, thereby mutually enhancing 3D perception in a resource-efficient manner.

Moreover, to further illustrate the practicality of our SDTR model in terms of inference speed, we have included FPS results in Table \ref{ablation1}. As observed, the computational burden for inference is marginally increased, due to the substantial DE branch only being present during the training stage.

\begin{figure*}[htp]
\centering 
   \includegraphics[width=1.0\textwidth]{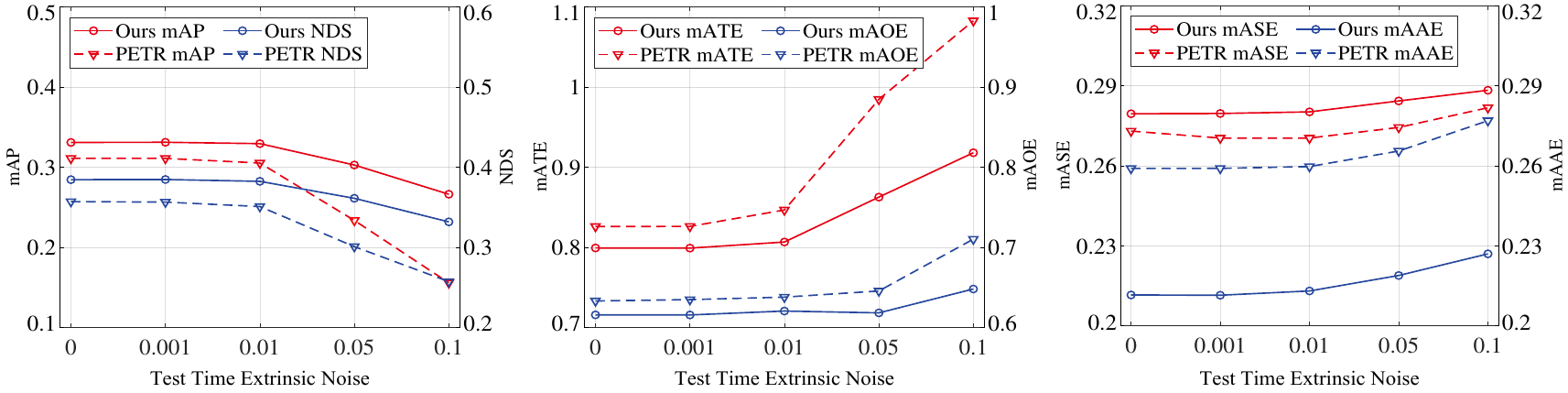}
   \caption{\textbf{The detection results on the nuScenes \textit{val} set with different extrinsics noises.} \sqre{In particular, random Gaussian noise is added to the camera extrinsic during testing. The findings suggest that our SDTR model exhibits superior robustness under large extrinsic noise.}}
   \label{noise}
\end{figure*}

\subsubsection{Combination of Losses} In practice, it has been observed that the depth loss typically exhibits significantly higher numerical values compared to those of the semantics loss. To determine the optimal ratio between their respective combination weights and achieve a balance between these two losses, we conducted a series of experiments, as presented in Table \ref{ablation2}. Based on empirical results, we find that setting the ratio of semantics loss to depth loss at 3 results in the highest accuracy. Further increasing the weight of semantics loss beyond this ratio does not yield a corresponding improvement in performance.

\sqre{\subsubsection{Results of Per-object Categories} We present Intersection over Union (IoU) metrics for all object categories in Table \ref{per_iou}, which serve to validate the effectiveness of our approach in segmenting objects, particularly small and distant objects. It's worth noting that a direct comparison with existing literature on per-object segmentation encounters methodology constraints, primarily because they adopt a train/validation split divergent from ours. In spite of this, our findings still demonstrates that the proposed SDTR maintains robust performance, notably in localizing small, distant categories.}

\begin{table}[t]
\caption{\sqre{\textbf{Ablation studies on small distant categories.} "Trans.", "Vel.", and "Attr." represent the translation, velocity, and attribute errors in 3D detection, respectively. We use the notation "w/o" to denote the model without the PQB module, and "w" to denote the model with the PQB module.}}
\centering
\resizebox{\columnwidth}{!}{
\begin{tabular}{r|cc|cc|cc}
\toprule
\multirow{2}{*}{Class} & \multicolumn{2}{c|}{Trans.↓} &  \multicolumn{2}{c|}{Vel.↓} & \multicolumn{2}{c}{Attr.↓} \\ 
                       & w/o          & w           & w/o         & w           & w/o         & w           \\  \midrule
Motor                  & 0.734        & 0.726        & 1.615       & 1.517       & 0.148       & 0.126       \\
Bike                   & 0.725        & 0.714        & 0.484       & 0.427       & 0.028       & 0.013       \\
Ped.                   & 0.727        & 0.722        & 0.799       & 0.781       & 0.352       & 0.298       \\ \bottomrule
\end{tabular}}
  \label{distant}
\end{table}

\subsubsection{Analysis on Multi-task Learning} We also investigate the impact of multi-task learning within the proposed framework by varying the loss weight assigned to each task.  As presented in Table \ref{multi_task}, our single-task model demonstrates higher accuracy across the majority of tasks, with the exception of vehicle segmentation. It is widely acknowledged that joint learning often yields a slightly lower precision than single-task learning \cite{Fifty2021EfficientlyIT, Crawshaw2020MultiTaskLW}. However, in the context of our study, it is possible that the improved precision in vehicle segmentation achieved through joint learning can be attributed to the 3D position information of vehicles acquired from 3D detection. Furthermore, our findings suggest that assigning a higher loss weight to the BEV segmentation task than the detection task in joint learning can lead to a performance increase of approximately 0.5\% to 1.2\%.

\sqre{\subsubsection{Analysis on the Prior-guided Query Builder.} To assess the effectiveness of our proposed PQB in detecting small distant objects, we present the results of our experiments on translation, velocity, and attribute errors across three distant classes using different model configurations: without PQB and with PQB. The findings, as illustrated in Table \ref{distant}, confirm the substantial improvement our PQB module contributes to the detection and accurate localization of these typically challenging distant objects. By visibly reducing false negatives, the PQB has proven instrumental in enhancing the overall precision of our model, particularly in scenarios traditionally prone to detection errors.}

\begin{figure}[t]
\centering
   \includegraphics[width=0.93\columnwidth]{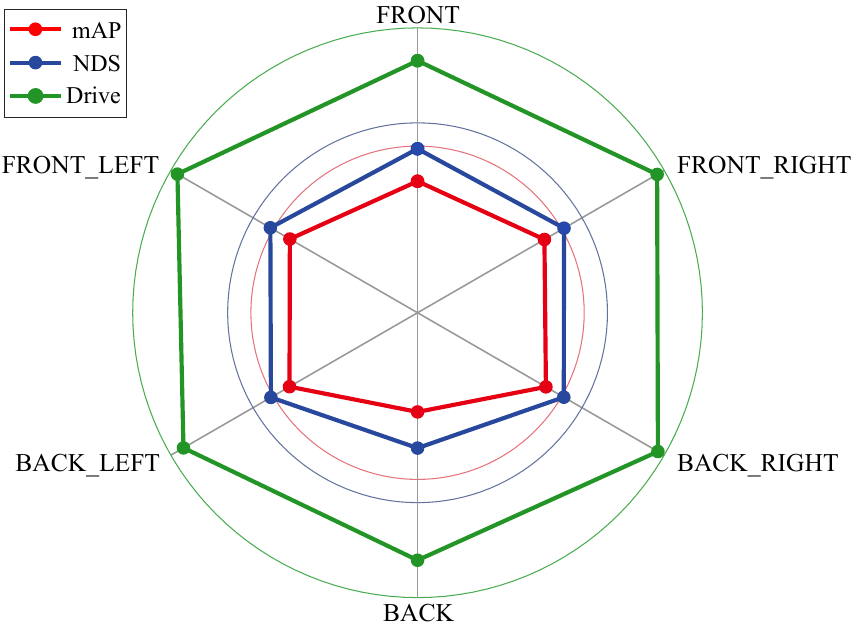}
   \caption{\textbf{The performance on the nuScenes \textit{val} set with camera drops.} For each metric, the points and the circle denote the results w/ and w/o camera drop. The closer to the center, the greater the degradation.}
   \label{camdrop}
\end{figure}

\subsubsection{Model Robustness}

\qy{The reliability of autonomous vehicles significantly hinges on the robustness of sensors. To evaluate the robustness of our model against potential sensor damage, we introduce two common types of sensor error during the testing phase.
Figure \ref{noise} illustrates a comparison of our SDTR model and the PETR model under various degrees of extrinsic noise in the matrix. The findings suggest that our SDTR model demonstrates superior robustness against extrinsic noise as compared to the PETR model, as observed across all performance metrics. 
Differing from PETR, which heavily relies on 3D coordinates computed by camera parameters, our framework leverages depth predictions to perceive precise 3D positions. In addition, the input-dependent queries in our model facilitate a comprehensive global perception that remains largely unaffected by camera parameters.}
Furthermore, we conduct a test where we randomly remove several camera images from each sample to evaluate the robustness of our model to camera dropout. As shown in Figure \ref{camdrop}, our SDTR model exhibits a relatively high level of accuracy even in the absence of training on such sensor errors. However, it is worth noting that the performance degradation caused by the loss of the BACK camera, which has a wider field of view, is more significant.

\section{Conclusion}
In this paper, we propose a unified framework, named SDTR, for addressing the challenges associated with multi-camera 3D object detection and BEV segmentation. In contrast to conventional techniques that rely on a tightly coupled learning process to extract categorical and 3D positional information, our approach first adopts S-D Encoders to explicitly extract semantic and depth priors. This decoupling enables greater flexibility and efficiency in our approach. Furthermore, we propose a Prior-guided Query Builder that transforms input-independent queries into input-aware ones. Experiments on the nuScenes and Lyft benchmarks demonstrate that our \sqre{SDTR} significantly improves the recovery of semantic information and 3D positions of objects, leading to better performance in both tasks.

\bibliographystyle{IEEEtran}
\bibliography{ieeeconf}

\vfill

\end{document}